\documentclass[conference]{IEEEtran}
\IEEEoverridecommandlockouts
\usepackage{cite}
\usepackage{amsmath,amssymb,amsfonts}
\usepackage{graphicx}
\usepackage{textcomp}
\usepackage{xcolor}
\def\BibTeX{{\rm B\kern-.05em{\sc i\kern-.025em b}\kern-.08em
    T\kern-.1667em\lower.7ex\hbox{E}\kern-.125emX}}

\usepackage[
    colorlinks=true,
    breaklinks=true,
    allcolors=blue
]{hyperref}

\usepackage[ruled,vlined]{algorithm2e}
\usepackage{algpseudocode}
\usepackage{multirow}
\usepackage{float}
\usepackage{booktabs}
\usepackage{subfig}

\begin{document}

\title{Privacy in Federated Learning with Spiking Neural Networks}

\author{\IEEEauthorblockN{Dogukan Aksu, Jesus Martinez del Rincon, Ihsen Alouani}
\IEEEauthorblockA{\textit{Centre for Secure Information Technologies (CSIT)} \\
\textit{Queen's University Belfast, UK}
}}


\maketitle

\begin{abstract}
Spiking neural networks (SNNs) have emerged as prominent candidates for embedded and edge AI. Their inherent low power consumption makes them far more efficient than conventional ANNs in scenarios where energy budgets are tightly constrained. In parallel, federated learning (FL) has become the prevailing training paradigm in such settings, enabling on-device learning while limiting the exposure of raw data. 
However, gradient inversion attacks represent a critical privacy threat in FL, where sensitive training data can be reconstructed directly from shared gradients. While this vulnerability has been widely investigated in conventional ANNs, its implications for SNNs remain largely unexplored. 
In this work, we present the first comprehensive empirical study of gradient leakage in SNNs across diverse data domains. SNNs are inherently non-differentiable and are typically trained using surrogate gradients, which we hypothesized would be less correlated with the original input and thus less informative from a privacy perspective.
To investigate this, we adapt different gradient leakage attacks to the spike domain.
Our experiments reveal a striking contrast with conventional ANNs: whereas ANN gradients reliably expose salient input content, SNN gradients yield noisy, temporally inconsistent reconstructions that fail to recover meaningful spatial or temporal structure. These results indicate that the combination of event-driven dynamics and surrogate-gradient training substantially reduces gradient informativeness.
To the best of our knowledge, this work provides the first systematic benchmark of gradient inversion attacks for spiking architectures, highlighting the inherent privacy-preserving potential of neuromorphic computation.
\end{abstract}


\section{Introduction}
\label{sec:intro}
The rapid deployment of machine learning (ML) models in privacy-sensitive domains such as healthcare, finance, and surveillance has raised growing concerns about the unintended leakage of private information through shared model parameters or gradients. Recent research has shown that model gradients, which are often exchanged during federated or distributed training, can be exploited by adversaries to reconstruct private training samples with alarming fidelity, a class of attacks known as gradient leakage or gradient inversion attacks~\cite{zhu2019deep, geiping2020inverting}. These findings highlight a critical vulnerability in collaborative and federated learning (FL) systems, where gradients are assumed to be benign communication artifacts.

\begin{figure}[tp]
  \centering
   \includegraphics[width=0.9\columnwidth]{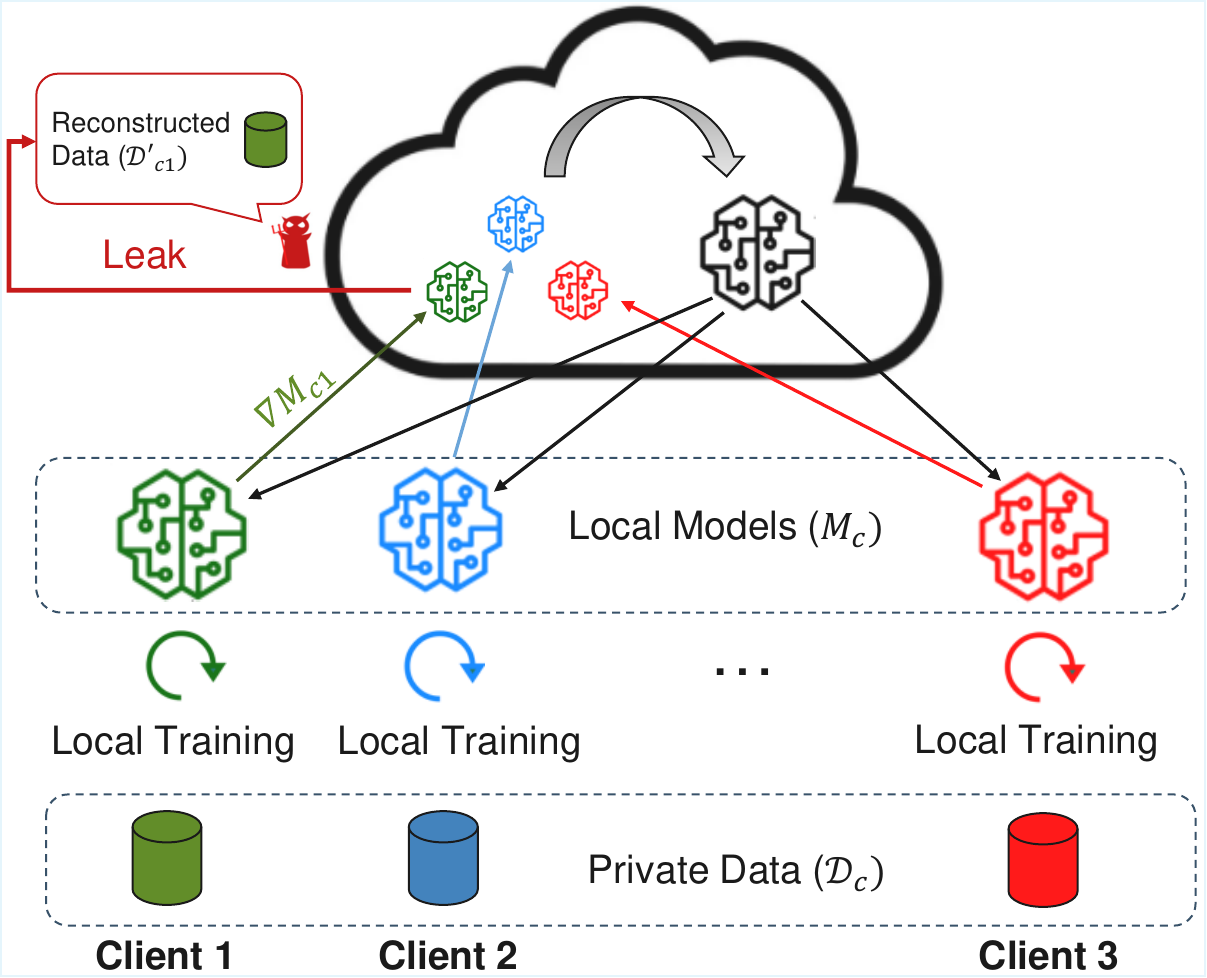}

   \caption{Gradient inversion attacks in FL.}
   \label{fig:leakage}
\end{figure}

Figure~\ref{fig:leakage} illustrates how this privacy breach occurs in FL. During local training, each client updates its model on private data and transmits gradients to a central server for aggregation. Although raw data never leave the client, these gradients encode sufficient information for adversaries to recover visual or semantic content of the original data. This vulnerability underscores the need for models that are inherently resistant to gradient-based inference. 

Spiking neural networks (SNNs), inspired by the discrete, event-driven signaling of biological neurons, have recently emerged as a promising class of models for low-power and neuromorphic computation~\cite{roy2019towards, tavanaei2019deep}. This low-power consortium makes them suitable for edgeAi appropriation, including FL systems based on edge clients. Moreover, by encoding information as temporally sparse spike trains rather than continuous activations, SNNs process dynamic sensory inputs efficiently while leveraging temporal structure in data such as speech, gesture, and vision streams~\cite{fang2021incorporating, rathi2021diet}. This inherently discrete and temporally extended computation paradigm raises a fundamental question: \textbf{Are SNNs more resilient to gradient leakage attacks than conventional Artificial Neural Networks (ANNs)?}

While gradient inversion has been extensively studied for ANNs, the privacy implications for SNNs remain largely unexplored. Unlike ANNs, the training of SNNs relies on surrogate gradients to approximate the non-differentiable spike function, and the forward dynamics unfold across multiple discrete timesteps. These characteristics disrupt the direct correspondence between input features and parameter gradients that gradient inversion exploits. Consequently, the temporal encoding mechanisms and the discontinuous activation functions inherent to SNNs may serve to inherently mitigate information leakage; however, this hypothesis remains to be empirically validated.

In this work, we present, to the best of our knowledge, the first systematic investigation of gradient leakage attacks on SNNs. We first adapt three canonical inversion methods, deep leakage from gradients (DLG)~\cite{zhu2019deep}, improved deep leakage from gradients (iDLG)~\cite{zhao2020idlg}, and generative regression neural network -- a data leakage attack for federated learning (GRNN)~\cite{yin2021see}, to the spiking domain, incorporating surrogate gradient computation and multi-step temporal propagation. Then, our hypothesis is evaluated across four public datasets and SNNs are directly compared both quantitatively and qualitatively against their ANN counterparts, demonstrating a significantly higher resilience against gradient leakage attacks.

The main contributions of this paper can be summarized as follows:
\begin{enumerate}
     \item \textbf{SNNs as gradient leakage preserving architectures:} We propose spiking and neuromorphic neural networks as privacy-preserving-by-design architectures, suitable for FL systems due to their inherent properties to prevent gradient leakage.
    \item \textbf{Adaptation of gradient inversion attacks to the spiking domain:} We adapt three state-of-art canonical inversion attacks, DLG, iDLG and GRNN, to handle the temporal encoding, surrogate gradient computation, and binary spike representation in SNNs. 
    \item \textbf{Empirical evidence of inherent privacy in spiking computation:} Through extensive experiments on four different datasets, including neuromorphic native datasets,  we demonstrate that SNNs significantly resist gradient leakage, producing noisy, temporally inconsistent, and semantically degraded reconstructions, highlighting the natural privacy-preserving properties of spiking dynamics.
\end{enumerate}

The implementation and experimental setup will be publicly available at \href{https://github.com/aksudogukan/Gradient-Leakage-SNN-Federated-Learning}{this repository}.

\section{Background and Related work}
\label{sec:related_works}

Gradient inversion, also known as gradient leakage, has emerged as a fundamental privacy threat in distributed and federated learning. 
The pioneering work of Zhu and Han \cite{zhu2019deep} demonstrated that shared gradients can reveal highly detailed training samples, initiating the study of Deep Leakage from Gradients (DLG). 
Subsequent improvements such as iDLG \cite{zhao2020idlg} and analytic label inference further enhanced reconstruction efficiency by estimating labels directly from gradient statistics. 
Later studies extended gradient inversion to stronger optimization setups \cite{geiping2020inverting, yin2021see, zhu2020r} and introduced generative priors or regularization terms to improve perceptual realism of recovered images. 
Other attack variants explored batch-wise gradient matching, partial gradient observation, or multi-round aggregation scenarios in federated learning \cite{hitaj2017deep, pasquini2021unleashing, geng2023improved}. 
These works collectively demonstrate that gradient sharing—even without access to model parameters or data—poses a concrete privacy risk. 
Beyond iterative matching, recent works proposed to learn direct mappings from gradients to input samples using neural generators. 
Such generative inversion methods \cite{jeon2021gradient, fang2024privacy} train an auxiliary network to reconstruct data from gradient features, enabling faster inference and generalization to unseen samples. 
While effective on high-resolution images, these approaches rely heavily on the strong spatial priors of natural image statistics and thus are less directly applicable to non-image modalities such as event-based spike streams. Their effectiveness on SNNs have not been tested till date.

Numerous defense strategies have been proposed to mitigate gradient leakage. 
Cryptographic approaches such as secure aggregation \cite{bonawitz2017practical} prevent raw gradient exposure, while statistical methods like Differentially Private SGD (DP-SGD) \cite{abadi2016deep} inject calibrated noise into gradient updates to ensure formal privacy guarantees. 
Heuristic techniques, including gradient pruning, compression, or obfuscation \cite{sun2021soteria, yue2023gradient}, can provide partial protection but are often vulnerable to adaptive attackers. 
Recent studies further suggest that architectural properties—such as non-linear activation saturation or stochasticity—may naturally reduce gradient–input correlation \cite{kaissis2021end, pasquini2021unleashing}. Some of these approaches have been ported to SNN architectures, such as Aitsam et al.\ \cite{aitsam2025differentially} where differential privacy were applied to SNNs, or \cite{kim2022privatesnn} where encryption during data conversion is applied. 
All these approaches, however, will result on a compromise on accuracy (DP) or computational load (encryption), which may not be necessary or could be reduced if there exist inherently robust neural models against gradient leakage.

Spiking Neural Networks (SNNs) process information through discrete events and exhibit temporal dynamics that differ fundamentally from traditional artificial neural networks (ANNs). 
Their training commonly relies on surrogate gradient methods \cite{neftci2019surrogate, wu2018spatio, fang2021deep}, which approximate the derivative of the non-differentiable spike function with smooth substitutes such as ATan or sigmoid functions. 
This surrogate-based backpropagation introduces approximation noise and temporal dependencies that potentially obfuscate direct gradient–input relationships. 

Only a few studies have begun to analyze the privacy implications of neuromorphic architectures. 
For instance, preliminary evidence suggests that spike discretization and temporal coding can reduce vulnerability to Membership Inference Attacks (MIA)\cite{moshruba2024neuromorphic,  guan2025privacy, moshruba2025privacy, li2024membership}. Beyond MIA, other privacy attacks such as model inversion  \cite{poursiami2024brainleaks} and data and class leakage \cite{kim2022privatesnn} during data modality conversion are explored.
Despite these initial insights on better privacy preservation on SNNs, gradient inversion attacks remain unexplored. 
Existing literature extensively studies gradient inversion in conventional ANNs, but its applicability to spiking architectures and event-based data remains under-examined and lacks systematic evaluation.




\section{Gradient Inversion on Spiking Architectures}
\label{sec:adapt_attack}
In this section, we extend three gradient inversion attacks to operate on SNNs:

\begin{itemize}
\item \textbf{DLG \cite{zhu2019deep}:} Dummy inputs $x'$ and dummy labels $y'$ are optimized jointly so that the gradients they produce $g'=\nabla_W \mathcal{L}_{dummy}$ closely match the victim gradients with respect to all network parameters, i.e. to minimize $\sum_{l}\| g'_l - \nabla W_l \|_2^2$ , where $l$ indicates a given layer.

\item \textbf{iDLG \cite{zhao2020idlg}:} Labels are first inferred directly from the last $L$ linear layer’s gradient,  $y' = \arg\min_{i}\;\sum_j \nabla W^{i}_{L,j}$, being $i$ a class index and $j$ the neuron index of the previous layer, 
thereby simplifying the optimization process to reconstruct only the input sample.

\item \textbf{GRNN \cite{ren2022grnn}:} Instead of directly optimizing dummy inputs, a generative neural network is trained to predict candidate inputs from observed gradients.
\end{itemize}

Our adaptation closely follows the canonical DLG, iDLG and GRNN formulations but incorporates temporal dynamics and surrogate gradient computation specific to spiking models. At each attack iteration, dummy inputs are propagated through the multi-step SNN to generate surrogate gradients, which are matched against the victim gradients. 

\subsection{Adaptation to the Spike Domain}

To adapt gradient inversion attacks to the spiking domain, both the input representation and gradient-matching loss to account for temporal dynamics needs to be adapted. To capture temporal dynamics, each input sample is represented over $T$ discrete simulation steps, forming an input tensor of shape $[T, B, C, H, W]$, where $(H,W)$ represent the input resolution for rows and columns, $C$ the number of channels and $B$ denotes the batch size. This data representation must be mimicked to create the dummy inputs $x'$. Dummy spikes are initialized as small non-negative probability distribution to reflect physical plausibility and sparse spike activity rather than pixel intensities.

\begin{equation}
    x' \sim |\mathcal{N}(0,\sigma)|^{T\times B\times C\times H\times W}
    \label{eq:ini}
\end{equation}

After initiating, during forward propagation, the network produces class-wise membrane potential outputs at each time step, which are aggregated by temporal averaging to obtain the final classification prediction. During optimization, Integrate-and-Fire (IF) spiking neurons utilize surrogate gradient functions to replace the non-differentiable spike function with a smooth approximation, thus enabling gradient-based optimization. The reconstructed gradients $\nabla_W'$ are computed by backpropagating through the surrogate gradient pathways of the SNN, and the objective minimizes the squared difference between the dummy and victim gradients, i.e.,

\begin{equation}
    \mathcal{L}_G = \sum_{l} \|\nabla W'^{(l)} - \nabla W^{(l)}\|_2^2
    \label{eq:grad-loss}
\end{equation}

\subsection{Threshold-based Binarization for Spiking Attacks}
The reconstructed spikes are constrained to remain positive throughout the process to preserve biological plausibility and prevent invalid activations, reflecting the binary (0 or 1) nature of spike-event representations. However, reconstructed dummy spikes from gradient inversion often take continuous values due to the use of surrogate gradient computation. To ensure correct spike-domain representation, we propose two thresholding strategies: Algorithm~\ref{alg:post-threshold} post-optimization thresholding (applied after attack) and Algorithm~\ref{alg:in-threshold} in-optimization thresholding (applied during backpropagation).

\begin{algorithm}
\caption{Post-Optimization Thresholding}
\label{alg:post-threshold}
\KwIn{Reconstructed dummy tensor $x'$, threshold $\tau$}
\KwOut{Binarized tensor $\tilde{x}'$}
\BlankLine
\ForEach{$t$ in timesteps}{
    $\tilde{x}'_t = \begin{cases}
      1, & x'_t > \tau \\
      0, & \text{otherwise}
    \end{cases}$ 
}
\end{algorithm}

\begin{algorithm}
\caption{In-Optimization Thresholding}
\label{alg:in-threshold}
\KwIn{Dummy tensor $x'$, threshold $\tau$, optimizer $O$, gradients $\nabla W$}
\KwOut{Binarized dummy $x'$ after each iteration}
\BlankLine
\For{$k \gets 1$ \KwTo $N$}{
    Compute surrogate loss $\mathcal{L}_G$ from Eq.~\eqref{eq:grad-loss}\;
    Update $x'$ using $O(\mathcal{L}_G)$\;
    \textbf{Binarize:} $x' \gets (x' > \tau)$ 
}
\end{algorithm}

Both strategies encourage spike-level consistency. The post-optimization threshold is simpler and decoupled from optimization dynamics, while the in-optimization variant enforces native spike-domain behavior during backpropagation, aligning reconstructed tensors with binary spiking activity.

\section{Experimental Methodology}
\begin{figure*}[t]
  \centering
   \includegraphics[scale=0.5]{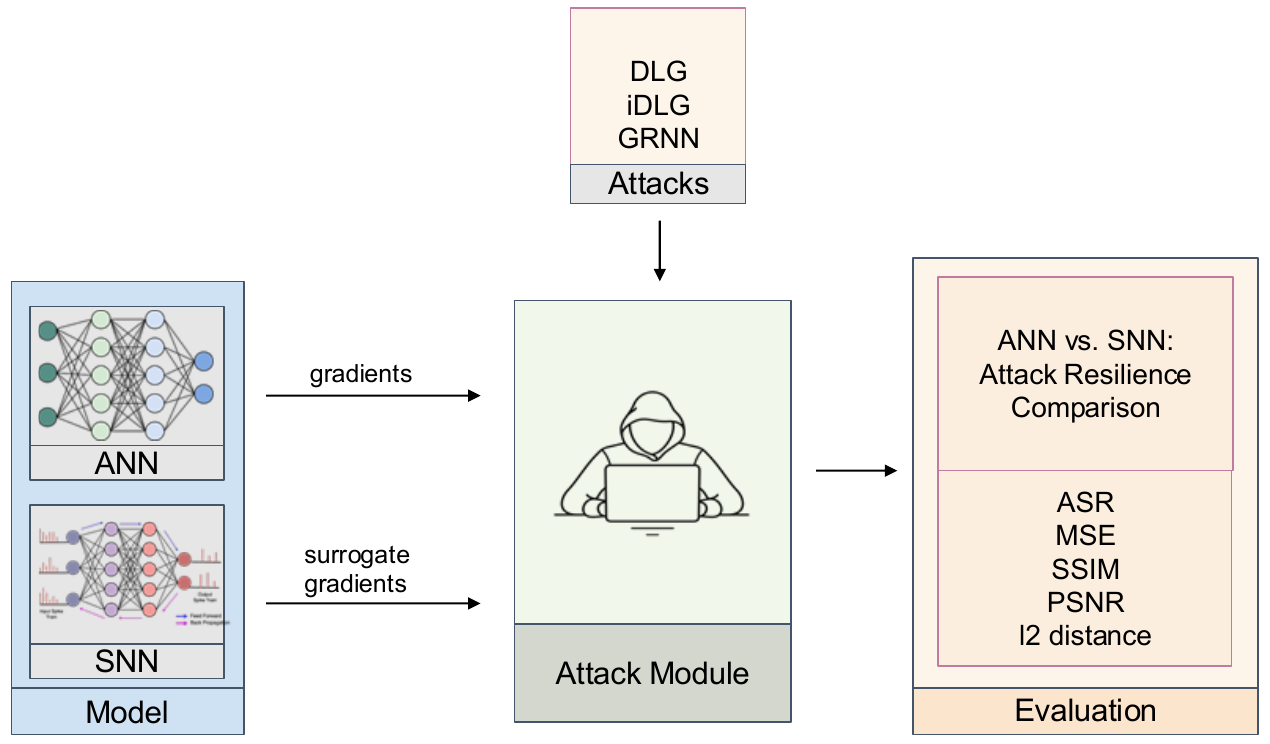}
   \caption{Overview on the experimental methodology.}
   \label{fig:eval}
\end{figure*}

In this section we describe the empirical methodology employed to evaluate the gradient leakage resilience of SNNs, whose surrogate gradient mechanism inherently obfuscates gradient information, thereby reducing the effectiveness of gradient leakage attacks without requiring external privacy-preserving techniques, such as differential privacy. The experimental overview is depicted in Figure~\ref{fig:eval}, to ensure a like-for-like comparison between ANNs and SNNs under a set of gradient leakage attacks, and its quantitative comparison using multiple metrics.

\noindent\textbf{Threat Model.} 
We consider a standard gradient inversion setting with an honest-but-curious server threat model, following prior works such as~\cite{zhu2019deep, zhao2020idlg}. The attacker can intercept unencrypted client gradients transmitted to the server (passive eavesdropper) but has no access to the victim’s private training data. They are assumed to know the victim model architecture (ANN or SNN), loss function, and general hyperparameters (e.g., batch size, number of timesteps \(T\)), enabling them to instantiate a local copy for gradient matching. 
During the gradient inversion process, the SNN model parameters remain fixed and unknown, and the attacker only has access to the gradients intercepted from the client-server communication that occurs during aggregation in the FL system.

\noindent\textbf{Attack set.} To evaluate the robustness against gradient leakage, we apply a range of widely used state-of-art gradient inversion attacks, DLG, iDLG, and GRNN to both ANNs and the SNN counterparts, by using the extended attacks as described in Section~\ref{sec:adapt_attack}. 

\noindent\textbf{Datasets.} We evaluate the gradient inversion attacks on four datasets: MNIST~\cite{li2025temporal}, CIFAR-100~\cite{sami2025gradient}, LFW~\cite{yang2025deep}, and DVS128 Gesture~\cite{amir2017low}. The first three represent static image data modality, while DVS128 Gesture introduces an event-based spiking input modality. This additional modality, while it will not allow us to compare ANNs on it, will enable us to verify if the observed gradient leakage robustness is intrinsic to SNNs or a subproduct of the additional spiking encoding required to process image data into SNNs.  
We employ the standard train and test splits of the MNIST, CIFAR~100, and DVS128 Gesture to train the corresponding judge models. For the LFW dataset, since it lacks predefined splits, we randomly assign 80\% of the data for training and 20\% for validation when training the ResNet-50 (initialized with IMAGENET1K\_V1 weights) combined with an ArcFace head over 5,749 identity classes. For the victim models, we follow the original gradient inversion setting, where attacks are applied directly on gradients computed from randomly selected samples within the training set, no additional data split or retraining is performed. 

For each dataset, the input resolution $(H,W)$ and channel configuration $C$ are adjusted accordingly, while the number of simulation timesteps $T$ is kept constant to ensure fair temporal comparison across SNN experiments.

\noindent\textbf{Neural network architectures.} 
For all image modality datasets, both LeNet and its spiking counterpart (SNN-LeNet) are used as victim models of the gradient inversion attacks to ensure a fair comparison between conventional and spiking representations. For consistency, SNN-LeNet is also use for the spike modality dataset. LeNet choice is justified by its selection by all the original gradient inversion attacks papers  \cite{zhu2019deep, zhao2020idlg, ren2022grnn}.

\noindent\textbf{Metrics and judge models.} Since gradient inversion attacks aim to reconstruct training samples given a corresponding captured gradient, their effectiveness can be measured by how recognizable is the reconstruction judged by an external observer. We propose to measure the reconstruction quality from the point of view of two different observer, a human observer focusing on the fidelity of the reconstructed sample -perceptibility- regarding the corresponding original sample, and an ML observer, focusing on the semantics, able to automatically recognise and/or reuse the sample as a particular class representative. The former is measured with a set of sample-to-sample similarity metrics such as Mean Square Error (MSE), Peak Signal-to-Noise Ratio (PSNR), Structural Similarity (SSIM) and $\ell_2$ distance, while the latter is measured using a pretrained classification model employed as a judge model, whose accuracy on the prediction of the corresponding class of the reconstructed sample is defined as the attack success rate (ASR) of fooling the judge model.

Specifically, a pre-trained  LeNet is used as the judge for MNIST and CIFAR-100 given the simplicity of the dataset. A more complex pre-trained ResNet-50 combined with an ArcFace embedding head is used for LFW to assess face identity, in par with the complexity of the dataset and task, where this Resnet architecture has reported effective \cite{ubayashi2010archface}. For DVS128 Gesture, a bespoke spiking convolutional network with residual connections serves as the judge to evaluate classification performance on reconstructed spike frames. The model adopts a hierarchical convolutional architecture composed of Leaky Integrate-and-Fire (LIF) neurons operating in a multi-step temporal setting. Each convolutional block includes residual shortcuts to enhance gradient flow across time and depth. Training is performed using surrogate gradient backpropagation with the ATan surrogate function, and temporal mean pooling is applied to the membrane potentials at the output layer to obtain class scores. The architecture and parameter configuration are summarized in Table~\ref{tab:dvs_snn_arch}. The judge models listed in Table~\ref{tab:judge_models} are used to evaluate the reconstructed samples and quantify the attack success across different datasets.

\begin{table}[tp]
\centering
\scriptsize
\caption{Architecture of the Residual Spiking Neural Network (DVSGestureSNN) used as judge to the evaluation model.}
\label{tab:dvs_snn_arch}
\renewcommand{\arraystretch}{1.1}
\begin{tabular}{p{2.3cm}p{4.8cm}}
\toprule
\textbf{Layer} & \textbf{Description} \\
\midrule
Input & Spike frames $\mathbf{x} \in \mathbb{R}^{T \times B \times 2 \times 128 \times 128}$ \\
Conv1 Block & Conv2d(2, 64, 3×3, stride=1, pad=1) → BN → LIF($\text{ATan}$) \\
Conv2 Block & Conv2d(64, 128, 3×3, stride=2, pad=1) → BN → LIF($\text{ATan}$) → ResidualBlock(128) \\
Conv3 Block & Conv2d(128, 256, 3×3, stride=2, pad=1) → BN → LIF($\text{ATan}$) → ResidualBlock(256) \\
Conv4 Block & Conv2d(256, 512, 3×3, stride=2, pad=1) → BN → LIF($\text{ATan}$) → ResidualBlock(512) \\
Pooling & Global average pooling over spatial dimensions \\
Classifier & Fully-connected layer: Linear(512, 11) \\
Temporal Aggregation & Time-averaged logits across $T$ timesteps \\
\bottomrule
\end{tabular}
\end{table}

\begin{table}[tp]
\centering
\scriptsize
\caption{Judge models and their corresponding test accuracies used for evaluating reconstructed samples across datasets.}
\label{tab:judge_models}
\renewcommand{\arraystretch}{1.1}
\begin{tabular}{p{1.4cm} p{2.6cm} c}
\toprule
\textbf{Dataset} & \textbf{Judge Model} & \textbf{Acc. (\%)} \\
\midrule
MNIST & LeNet & 97.75 \\
CIFAR-100 & LeNet & 24.62 \\
LFW & ResNet-50 (weights with "IMAGENET1K\_V1") + ArcFace head & 100 \\
DVS128 Gesture & DVSGestureSNN & 92.80 \\
\bottomrule
\end{tabular}
\end{table}

\subsection{Implementation Considerations}

Our implementation employs a multi-step LeNet-inspired SNN architecture, called LeNet5SNN, constructed using the SpikingJelly \footnote{\url{https://spikingjelly.readthedocs.io/zh-cn/latest/activation_based_en/basic_concept.html}} framework in its activation-based mode. 

In all DLG and iDLG experiments, optimization to minimize the gradient matching loss $L_G$ is performed using the L-BFGS optimizer\footnote{\url{https://pytorch.org/docs/stable/generated/torch.optim.LBFGS.html}}, which provides stable convergence for high-dimensional gradient matching problems. For GRNN, instead of iterative optimization, a regression network $R$ is trained to learn a mapping from gradients to candidate inputs, enabling generalization across multiple gradient samples while following the same reconstruction pipeline used for ANN and SNN backbones.

For image datasets, dummy inputs required by the gradient inversion attacks are initialized as near-uniform grayscale images  $\mathcal{U}(0.45,0.55)$ and temporally replicated across $T$ time steps. For the native spiking modality a Normal distribution, as defined by equation~\ref{eq:ini} with $sigma=0.1$ is used. Spike samples are composed of asynchronous event streams that are converted into frame sequences using the Tonic framework \footnote{\url{https://tonic.readthedocs.io/en/latest/index.html}} with the ToFrame transformation. 
A summary of the required implementation changes across modalities is provided in Table~\ref{tab:dvs_comparison}.

\begin{table}[H]
\centering
\footnotesize
\setlength{\tabcolsep}{3pt}  
\caption{Comparison of image and event-based DLG/iDLG implementations.}
\label{tab:dvs_comparison}
\renewcommand{\arraystretch}{1.1}
\begin{tabular}{p{1.6cm}p{2.9cm}p{3.2cm}}
\toprule
\textbf{Aspect} & \textbf{Image-modality} & \textbf{Spike-Modality} \\
\midrule
Data Type & Images & Event-based spike frames \\
Framework & PyTorch (vision) & Tonic + SpikingJelly \\
Input Format & $[B,C,H,W]$ & $[T,B,C,H,W]$ \\
Encoding & Frame replication & Event-to-frame \\
Model & LeNet/LeNetSNN & LeNetSNN \\
Dummy Init. & Uniform gray $\mathcal{U}(0.45,0.55)$ & Small positive spikes $|\mathcal{N}(0,0.1)|$ \\
Clamping & Pixel range $[0,1]$ & None (spike intensity) \\
Loss Func. & Cross Entropy (iDLG) / soft-label (DLG) & Same \\
Optimization & LBFGS ($\eta$) & Same \\
\bottomrule
\end{tabular}
\end{table}

For SNN ingestion, after preprocessing, each sample is represented as a spike tensor of shape $[T, C, H, W]$, where $T{=}20$ denotes the number of temporal bins, $C{=}2$ the polarity channels, and $(H,W){=}(128,128)$ the spatial resolution. 

\section{Experimental results}




\subsection{Experiment 1: Robustness comparison on image modality}

In this experiment we perform a like-by-like comparison of the three state-of-art gradient inversion attacks on the ANN and corresponding SNN counterpart on the three image datasets. The purpose is to highlight our hypothesis gradient leakage robustness of the spiking neural networks.

Following \cite{zhu2019deep, zhao2020idlg}, each ANN/SNN is randomly initialized prior to the attack. Gradients are computed on a randomly selected ground-truth sample set using this untrained model, and the attacker then optimizes dummy inputs to match these gradients. A thousand samples are used in this attack set. The model parameters remain fixed throughout the attack and are not updated. The reconstructed samples are then evaluated by the corresponding judge model and similarity metrics. 
 
\subsubsection{Quantitative evaluation}
Table~\ref{tab:snn_datasets} compares the inversion performance between LeNet and SNN-LeNet across datasets and attack types. In all settings, reconstructed samples from SNNs exhibit significantly lower ASR and perceptual fidelity compared to their ANN counterparts. This degradation reflects the effect of temporal encoding and surrogate gradient approximation, which obscure the direct correspondence between weights and input features. On average, SNN-LeNet exhibits a $\approx$ 49\% decrease in ASR compared to LeNet across all datasets and attack types. For instance, iDLG attack achieves an ASR of 97.3\% on LFW with high perceptual scores (SSIM $=0.91$), while SNN-LeNet yields only 32.2\% ASR and a 0.28 SSIM. Similar patterns are observed for MNIST and CIFAR-100. 

\begin{table}[H]
\centering
\caption{Comparison of gradient inversion performance on LeNet and SNN-LeNet across datasets. 
\#samples:1000.}
\label{tab:snn_datasets}
\resizebox{\linewidth}{!}{%
\begin{tabular}{l l l c c c c}
\toprule
\midrule
Attack & Dataset & Model & ASR (\%) & MSE~($\downarrow$) & SSIM~($\uparrow$) & PSNR~($\uparrow$ db) \\
\midrule
\multirow{6}{*}{DLG} 
  & \multirow{2}{*}{MNIST}     & LeNet     & 79.40 & 0.07 & 0.77 & 29.94 \\
  &                            & SNN-LeNet & 37.20 & 0.15 & 0.39 & 9.48 \\
  \cmidrule(lr){2-7}
  & \multirow{2}{*}{CIFAR-100} & LeNet     & 68.80 & 0.05 & 0.67 & 30.29 \\
  &                            & SNN-LeNet & 24.40 & 0.14 & 0.11 & 9.46 \\
  \cmidrule(lr){2-7}
  & \multirow{2}{*}{LFW}       & LeNet     & 92.50 & 0.05 & 0.57 & 25.34 \\
  &                            & SNN-LeNet & 83.80 & 0.16 & 0.05 & 8.06 \\
\midrule
\multirow{6}{*}{iDLG} 
  & \multirow{2}{*}{MNIST}     & LeNet     & 88.30 & 0.04 & 0.86 & 28.03 \\
  &                            & SNN-LeNet & 37.80 & 0.11 & 0.41 & 9.99 \\
  \cmidrule(lr){2-7}
  & \multirow{2}{*}{CIFAR-100} & LeNet     & 82.60 & 0.03 & 0.81 & 33.77 \\
  &                            & SNN-LeNet & 22.80 & 0.08 & 0.18 & 11.85 \\
  \cmidrule(lr){2-7}
  & \multirow{2}{*}{LFW}       & LeNet     & 97.30 & 0.01 & 0.91 & 33.95 \\
  &                            & SNN-LeNet & 32.20 & 0.07 & 0.28 & 11.89 \\
\midrule
\multirow{6}{*}{GRNN} 
  & \multirow{2}{*}{MNIST}     & LeNet     & 85.80 & 0.02 & 0.69 & 19.36 \\
  &                            & SNN-LeNet & 9.60 & 0.21 & 0.00 & 6.78 \\
  \cmidrule(lr){2-7}
  & \multirow{2}{*}{CIFAR-100} & LeNet     & 69.30 & 0.01 & 0.66 & 19.48 \\
  &                            & SNN-LeNet & 10.70 & 0.11 & 0.07 & 10.16 \\
  \cmidrule(lr){2-7}
  & \multirow{2}{*}{LFW}       & LeNet     & 88.10 & 0.01 & 0.66 & 18.85 \\
  &                            & SNN-LeNet & 55.10 & 0.11 & 0.39 & 9.66 \\
\midrule

\bottomrule
\end{tabular}%
}
\end{table}

\subsubsection{Qualitative evaluation}

\begin{figure}[tp]
    \centering
    \subfloat[\scriptsize{MNIST -- GT}]{
        \includegraphics[width=0.18\linewidth]{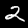}
    }
    \subfloat[\scriptsize{DLG(ANN)}]{
        \includegraphics[width=0.18\linewidth]{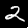}
    }
    \subfloat[\scriptsize{DLG(SNN)}]{
        \includegraphics[width=0.18\linewidth]{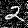}
    }
    \subfloat[\scriptsize{iDLG(ANN)}]{
        \includegraphics[width=0.18\linewidth]{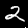}
    }
    \subfloat[\scriptsize{iDLG(SNN)}]{
        \includegraphics[width=0.18\linewidth]{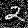}
    }\\[4pt]
    \subfloat[\scriptsize{CIFAR100-GT}]{
        \includegraphics[width=0.18\linewidth]{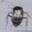}
    }
    \subfloat[\scriptsize{DLG(ANN)}]{
        \includegraphics[width=0.18\linewidth]{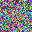}
    }
    \subfloat[\scriptsize{DLG(SNN)}]{
        \includegraphics[width=0.18\linewidth]{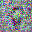}
    }
    \subfloat[\scriptsize{iDLG(ANN)}]{
        \includegraphics[width=0.18\linewidth]{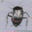}
    }
    \subfloat[\scriptsize{iDLG(SNN)}]{
        \includegraphics[width=0.18\linewidth]{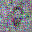}
    }\\
    \subfloat[\scriptsize{LFW -- GT}]{
        \includegraphics[width=0.18\linewidth]{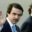}
    }
    \subfloat[\scriptsize{DLG(ANN)}]{
        \includegraphics[width=0.18\linewidth]{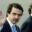}
    }
    \subfloat[\scriptsize{DLG(SNN)}]{
        \includegraphics[width=0.18\linewidth]{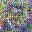}
    }
    \subfloat[\scriptsize{iDLG(ANN)}]{
        \includegraphics[width=0.18\linewidth]{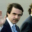}
    }
    \subfloat[\scriptsize{iDLG(SNN)}]{
        \includegraphics[width=0.18\linewidth]{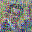}
    }
    \\

    \caption{Comparison of Ground Truth (GT) and reconstructed samples under DLG and iDLG attacks for ANN (LeNet5) and SNN (LeNet5SNN). 
    }
    \label{fig:recon_comparison_all}
\end{figure}

\begin{figure}[tp]
  \centering
  \setlength{\tabcolsep}{2pt}
  \renewcommand{\arraystretch}{0.8}

  \begin{minipage}{\linewidth}
    \centering
    \begin{tabular}{ccc}
      \includegraphics[width=0.22\linewidth]{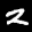} &
      \includegraphics[width=0.22\linewidth]{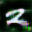} &
      \includegraphics[width=0.22\linewidth]{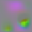} \\
      \scriptsize (i) MNIST-GT & \scriptsize (ii) ANN & \scriptsize (iii) SNN
    \end{tabular}\\
    {\scriptsize (a) MNIST}
  \end{minipage}

  \begin{minipage}{\linewidth}
    \centering
    \begin{tabular}{ccc}
      \includegraphics[width=0.22\linewidth]{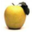} &
      \includegraphics[width=0.22\linewidth]{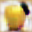} &
      \includegraphics[width=0.22\linewidth]{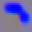} \\
      \scriptsize (i) CIFAR100-GT & \scriptsize (ii) ANN & \scriptsize (iii) SNN
    \end{tabular}\\
    {\scriptsize (b) CIFAR-100}
  \end{minipage}

  \begin{minipage}{\linewidth}
    \centering
    \begin{tabular}{ccc}
      \includegraphics[width=0.22\linewidth]{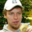} &
      \includegraphics[width=0.22\linewidth]{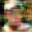} &
      \includegraphics[width=0.22\linewidth]{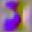} \\
      \scriptsize (i) LFW-GT & \scriptsize (ii) ANN & \scriptsize (iii) SNN
    \end{tabular}\\
    {\scriptsize (c) LFW}
  \end{minipage}

  \caption{
      Comparison of GRNN-based gradient inversion across datasets and model types. 
  }
  \label{fig:grnn_all}
\end{figure}

Figures~\ref{fig:recon_comparison_all} and~\ref{fig:grnn_all} demonstrate the comparison of ground truth (GT) images with their corresponding reconstructions under DLG/iDLG and GRNN attack methods respectively for both ANNs and SNNs. 
In almost all the cases, attack reconstructions on ANNs successfully recover the original samples with high fidelity, confirming the strong alignment between the gradient and the original input. As exception we depict a case where DLG collapses due to gradient explosion on a CIFAR100 sample, an effect already exhibit in the original paper \cite{zhu2019deep}. In contrast, SNN-based reconstructions appear to be highly noisy, particularly when reconstructing complex natural images from gradients is substantially more challenging. This qualitative results confirm the results in the previous section showing a significantly worse reconstruction when performing gradient inversion attacks on SNNs. Regarding attack types, SNNs shows a greater robustness against GRNN attack, where the reconstructed image does not seem to improve from the initial iterations attempt. 

\subsubsection{Analysis}
Across all datasets, gradient inversions on ANNs recover coarse visual semantics and preserve class identity to a reasonable degree, enabling approximate visual recognition of the original input. In contrast, SNNs exhibit greater robustness against such inversion attacks.  Of particular note are GRNN reconstructions for SNN, which appear heavily distorted, non-meaningful, and lack any recognizable structure or semantic content. 

Overall, these findings confirm that temporal dynamics and surrogate gradient computation in SNNs inherently mitigate gradient leakage risks. The non-differentiable spike function and multi-step temporal propagation effectively obfuscate the gradient-input correlation exploited by inversion attacks, resulting in reconstructions that are perceptually degraded and semantically ambiguous.

\subsection{Experiment 2: Gradient inversion on spiking modality}

To further validate the inherent robustness of spiking representations against gradient inversion, in this experiment we perform an additional evaluation on DVS128 Gesture dataset, whose data in natively produces as event-based spike trains. It could be argued that the previous improved robustness in SNNs vs ANNs may be partially due to an additional encoding to transform pixel data into asynchronous spike tensors. However, if similar ASR is achieved in this experiment where no additional encoding is done in the input data, this will reinforce the inherent robustness of SNN due to the non-differentiable functions and surrogate gradients. On the contrary, given the data modality, this experiment does not allow for ANN architectures to be tested.

Results are summarized in Table~\ref{tab:snn_dvsgesture}, including the effect of the threshold value used to covert the SNN output into a spike train $\in {0,1}$ .
Across all tested threshold values and strategies, both DLG and iDLG yield near-random ASRs between 7\% and 10\%  (DVS Gesture containn 11 classes yielding  random results at 9.1\%), with only marginal improvements at higher thresholds (e.g., iDLG reaching 10.64\% at $\tau{=}0.9$). Overall, these findings indicate that SNN effectively obfuscate input–gradient correlations, rendering DLG and iDLG attacks largely ineffective on event-based data.

\begin{table}[tp]
\centering
\scriptsize
\caption{Gradient inversion performance on the DVS128 Gesture dataset.}
\label{tab:snn_dvsgesture}
\begin{tabular}{c l c c c c}
\toprule
\midrule
{}&{}& \multicolumn{2}{c}{Post-Optimization $\tau$} & \multicolumn{2}{c}{In-Optimization $\tau$}\\
\midrule
Threshold ($\tau$) & Attack & ASR. (\%) & $\ell_2$ distance & ASR. (\%) & $\ell_2$ distance \\
\midrule
\multirow{2}{*}{0.1} 
  & DLG  & 9.41  & 504.03 & 9.60 & 457.2643  \\
  & iDLG & 9.19  & 508.09 & 9.60 & 457.2764 \\
\midrule
\multirow{2}{*}{0.25} 
  & DLG  & 7.67 & 257.75   & 7.10 & 100.1694  \\
  & iDLG & 8.54 & 270.65 & 7.80 & 100.1893\\
\midrule
\multirow{2}{*}{0.5} 
  & DLG  & 8.04 & 127.36 & 8.50 & 42.5694 \\
  & iDLG & 8.29 & 130.44 & 8.50 & 42.5694 \\
\midrule
\multirow{2}{*}{0.75} 
  & DLG  & 9.41 & 87.52 & 8.50 & 42.5647\\
  & iDLG & 9.65 & 85.45 & 8.50 & 42.5647  \\
\midrule
\multirow{2}{*}{0.9} 
  & DLG  & 8.54  & 75.47 & 8.50  & 42.5647\\
  & iDLG & 10.64 & 72.33  & 8.50  & 42.5647 \\
\midrule
\multirow{2}{*}{0.95} 
  & DLG  & 8.42  & 72.50  & 8.50  & 42.5647   \\
  & iDLG & 10.52 & 69.21  & 8.50  & 42.5647   \\
\midrule
\bottomrule
\end{tabular}%
\end{table}

\subsubsection{Spike reconstructed fidelity}
As similarity metric, $\ell_2$ is used instead of MSE/PSNR/SSIM for being more adequate to compare sparse binary sequences. Increasing the threshold reduces the overall $\ell_2$ distance—indicating suppression of spurious spike activations—but this improvement does not translate into better semantic reconstructions.

To contextualize these similarity results, Table~\ref{tab:ref_l2_stats} reports the natural intra- and inter-class $\ell_2$ variability within the original DVS128 Gesture dataset. The mean intra-class distance of $60.34$ and inter-class distance of $60.80$ suggest that samples from the same gesture class already exhibit high spatiotemporal diversity. The reconstructed samples reach comparable $\ell_2$ values ($\approx70-85$ under post-optimisation thresholding) or lower ($\approx$42.6 under a suboptimal in-optimization thresholding, see section~\ref{sec:abla}) only under aggressive thresholding, yet they still fail to convey coherent gesture patterns. In other settings, distances far exceed 100, confirming that optimization fails to recover any temporally structured spike activity. 

\begin{table}[H]
\centering
\scriptsize
\caption{Reference $\ell_2$ distance statistics computed on DVS128 Gesture. 
Intra-class measures distances between samples of the same gesture, while inter-class measures distances between samples of different gestures. Values are averaged over all 11 classes. 
}
\label{tab:ref_l2_stats}
\begin{tabular}{lcccc}
\toprule
\midrule
Type & Mean & Std. & Min & Max \\
\midrule
Intra-class  & 60.34 & 13.80 & 0.00  & 106.64 \\
Inter-class  & 60.80 & 11.51 & 14.93 & 105.15 \\
\midrule
\bottomrule
\end{tabular}
\end{table}

\subsubsection{Ablation Experiment: Thresholding strategy}
\label{sec:abla}

Our evaluation also compares our two proposed thresholding strategies: performing thresholding either after the attack optimization process (Algorithm~\ref{alg:post-threshold}) or during backpropagation, directly within the optimization loop such that dummy inputs are binarized at each iteration (Algorithm~\ref{alg:in-threshold}). 


Results in Table~\ref{tab:snn_dvsgesture} shows very similar results for both strategies in terms of ASR and semantic reconstruction. 
As main difference, the lower $\ell_2$ values of in-optimization thresholding arise from the stronger binarization constraint, which continuously limits the range of dummy activations rather than capturing meaningful spatiotemporal patterns. Consequently, enforcing discrete spikes during optimization does not enhance the attack’s ability to infer gesture structure—instead, it further restricts gradient expressiveness and reinforces the resilience of spiking representations against inversion-based leakage.

\subsubsection{Ablation Experiment: Threshold value}
\begin{figure}[tp]
  \centering
   \includegraphics[scale=0.37]{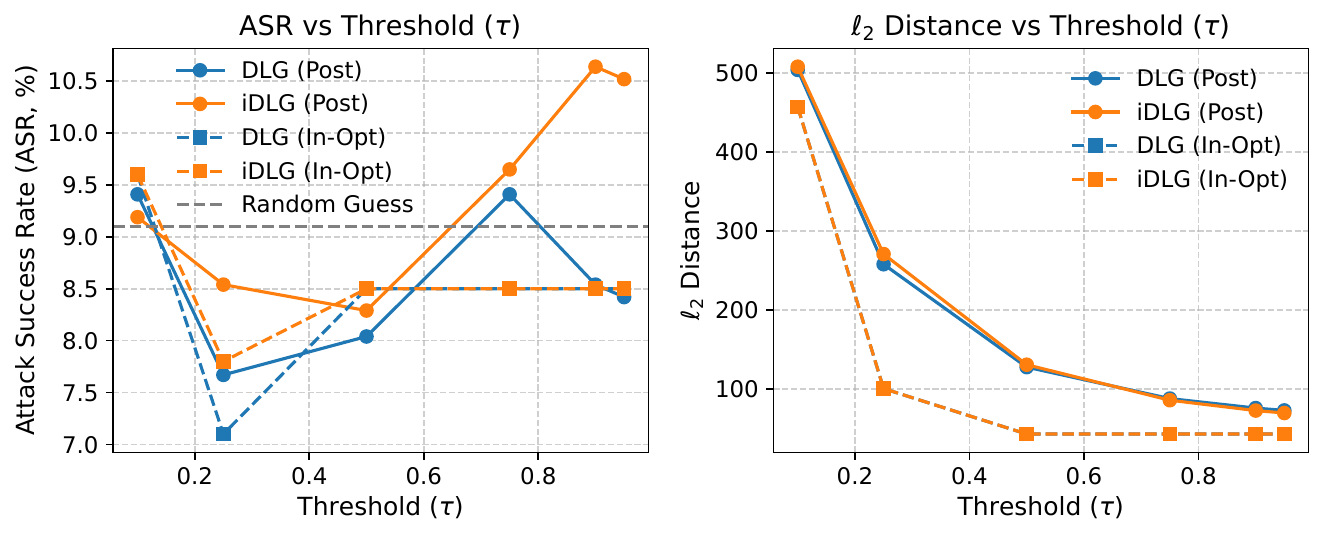}
   \caption{ASR (left) and $\ell_2$ distance (right)  for DLG and iDLG on DVS128 Gesture. 
  }
   \label{fig:accl2}
\end{figure}

Figure \ref{fig:accl2} illustrates the effect of applying different spike binarization thresholds on the DLG and iDLG attacks for the DVS128 Gesture dataset. As the threshold increases from 0.1 to 0.95, the $\ell_2$ distance between reconstructed and original spike tensors decreases sharply, indicating that higher thresholds yield sparser yet more stable reconstructions. However, the ASR (classification accuracy of the reconstructed samples) remains nearly constant and close or below to random-guess performance (around 9\% for 11 gesture classes), confirming that neither DLG nor iDLG successfully recovers meaningful spiking activity patterns. The minimal variation in ASR across thresholds demonstrates the robustness of SNN-based representations against gradient inversion attacks—binarized spike dynamics further obscure gradient-to-input correlations, resulting in reconstructions that lack discriminative gesture information.

\section{Concluding Remarks}
This paper presents the first comprehensive study of gradient inversion attacks on spiking neural networks (SNNs), extending the well-known DLG and iDLG frameworks to the temporal and event-driven domain and systematically comparing the effect of the attacks on SNN vs ANN counterparts. We proposed SNN-compatible inversion attacks that incorporate surrogate gradients, multi-step membrane dynamics, and temporal input encoding, enabling consistent evaluation across both conventional and spiking architectures. Through extensive experiments on image (MNIST, CIFAR-100, LFW) and event-based (DVS128 Gesture) datasets, we demonstrated that SNNs exhibit a strong inherent resistance to gradient leakage, with
attack success rates often dropping below roughly 40\% of ANN equivalents.

Our results show that while ANN-based models can be effectively inverted—often reconstructing high-fidelity images with accurate semantic content—spiking models yield noisy, temporally inconsistent reconstructions that fail to recover meaningful visual or event structure. This degradation arises from the discrete and temporally coupled nature of SNN activations, which disrupts the direct correspondence between gradients and input features. and, particularly, to the non-differentiable nature of spiking computation inherently reduces gradient information leakage. 

Overall, these findings highlight an important privacy-preserving property of spiking computation. 
Future work will explore formal privacy guarantees for SNNs and how their inherent robustness may relax the noise budgets required by privacy-preserving approaches such as differential privacy. 

\bibliographystyle{IEEEtran}
\bibliography{main}

\end{document}